\pdfoutput=1
\documentclass[runningheads]{llncs}
\usepackage{graphicx}
\usepackage{amsmath,amssymb} % define this before the line numbering.
\usepackage{tabularx}
\usepackage[nocompress]{cite}
\usepackage{algorithm}% http://ctan.org/pkg/algorithm
\usepackage{algpseudocode}% http://ctan.org/pkg/algorithmicx
\usepackage{color, colortbl}
\usepackage{dsfont}
\usepackage{amssymb}
\definecolor{Gray}{gray}{0.9}
\usepackage{hyperref}

\usepackage{times}
\usepackage{epsfig}
\usepackage{subcaption}
\usepackage{caption}

\newcolumntype{Y}{>{\hsize=.7\hsize\centering\arraybackslash}X}

\begin{document}
\title{Turning a Blind Eye: Explicit Removal of Biases and Variation from Deep Neural Network Embeddings} 

\titlerunning{Turning a Blind Eye}

\author{Mohsan Alvi\inst{1} \and
Andrew Zisserman\inst{1} \and
Christoffer Nell{\aa}ker\inst{2,3}}

\authorrunning{M. Alvi, A. Zisserman, and C. Nell{\aa}ker}

\institute{Visual Geometry Group, Department of Engineering Science, University of Oxford \and
Nuffield Department of Obstetrics and Gynaecology, University of Oxford \and
Big Data Institute, University of Oxford}
\maketitle              % typeset the header of the contribution
\begin{abstract}

Neural networks achieve the state-of-the-art in  image classification 
tasks. However, they can encode spurious variations or biases  that may be present in the training data. 
For example, training an age predictor on a  dataset that is not 
balanced for gender can lead to gender biased predicitons (e.g. wrongly predicting that males are older if only elderly males are in the 
training set). 

We present two distinct contributions: 

1) An algorithm that can remove multiple sources of variation from the 
feature 
representation of a network. We demonstrate that this algorithm can be 
used to remove 
biases from the feature representation, and thereby improve classification 
accuracies,  when training networks on extremely biased datasets. 

2) An ancestral origin database of 14,000 images of individuals from 
East Asia, 
the Indian subcontinent, sub-Saharan Africa, and Western Europe. 

We demonstrate on this dataset, for a number of facial attribute 
classification tasks, that we are able to remove racial biases from 
the network feature representation. 

\end{abstract}
%
% ------------------------------------------------------------
\section{\label{sec:intro}Introduction}
% ------------------------------------------------------------
The current state-of-the-art image recognition algorithms are based on convolutional neural networks \cite{he_deep_2015, krizhevsky_imagenet_2012, simonyan_very_2014}. These networks rely on large datasets of labeled images, to simultaneously generate a feature representation and a decision framework. This approach removes the need to handcraft features for any given problem but also gives rise to the question as to what feature representation the network has actually learned.

These models are trained on large datasets that contain a number of biases \cite{tommasi_deeper_2015, torralba_unbiased_2011} or spurious variations, that are irrelevant, or even problematic, for a given task (e.g. discriminating by gender or ancestral origin).
One such example is face recognition. Large publicly available face datasets are often composed of celebrities, such as \cite{cao_vggface2, msceleb1m, LFWTech, megaface, kostinger_annotated_2011, parkhi_deep_2015}. 
These can contain age, gender, and ancestral origin biases: for example, female celebrities tend to be younger than their male counterparts. This bias does not represent the real world outside of the movie business, reducing the usefulness of models trained on these datasets.
Furthermore, in cases where large datasets are not available, training algorithms are initialized from networks that have been trained on similar tasks, for which more data is available \cite{girshick_rich_2013, long_learning_2015, oquab_learning_2014}. This method, called fine-tuning, carries two potential issues: 1) spurious variations are learned from the small dataset, and 2) inheriting biases present in the large dataset.

The use of big data to train AI-models has, to name a few, been adopted by government agencies, institutions of law, human resources, and medical decision systems. A number of such models have been shown to make decisions based on the gender or ancestral origin of an individual, leading to concerns about their ``fairness'' \cite{pmlr-v81-buolamwini18a, ras_explanation_2018, sweeney_discrimination_2013}. With the recent enforcement of the General Data Protection Regulation laws\footnote{\url{https://www.eugdpr.org}}, individuals have the right to know the rationale behind an automated decision concerning them. To continue the adoption of deep learning, more needs to be done to make neural networks more transparent. One way to approach this issue is to prevent models from making decisions for the wrong reasons. 

To be sure that decisions are not being made due to biases, we must look beyond using accuracy as our only performance metric. An experiment by Zhao et al \cite{zhao_men_2017} showed that neural networks learn and amplify biases in the training data. Women were more often depicted in kitchens than men, so the network learned that being in the kitchen was a key feature for the identification of women. Though this may have been true for the dataset the classifier was trained on, in general, this is not a reliable indicator of the presence of a woman.

In this paper, we introduce an algorithm, inspired by a domain and task adaptation approach \cite{tzeng_simultaneous_2015}, that can be used to 1) to ensure that a network is blind to a known bias in the dataset, 2) improve the classification performance when faced with an extreme bias, and 3) remove multiple spurious variations from the feature representation of a primary classification task. We use age, gender, ancestral origin, and pose information for facial images to demonstrate our framework. 

As discrimination by ancestral origin is a type of spurious variation in many tasks, we have created a new labeled ancestral origin dataset, ``Labeled Ancestral Origin Faces in the Wild (LAOFIW)'', from publicly available images. This dataset contains 14,000 images of individuals whose ancestral origin is sub-Saharan Africa, Indian Subcontinent, Europe, and East Asia, with a variety of poses, illumination, facial expressions.

The rest of this paper is organized as follows. Section \ref{sec:data} introduces our LAOFIW dataset and other datasets we used for our experiments. Section \ref{sec:methods} discusses the methods to remove spurious variations from the feature representation of a network. In section \ref{subsec:expremovebias}, we present an experiment to remove a bias from the feature representation network. In section \ref{subsec:expremovepbias}, we investigate how removing an extreme bias can improve classification performance, and in section \ref{subsec:expremovemulti}, we investigate removing multiple spurious variations from a network. The results are detailed in section \ref{sec:results}. Finally, section \ref{sec:conclusion} summarises our findings.

%-------------------------------------------------------------
\section{\label{sec:relatedwork}Related Work}
% ------------------------------------------------------------

Image datasets are known to contain biases that cause models to generalize poorly to new, unseen data \cite{tommasi_deeper_2015, torralba_unbiased_2011}. This has been addressed by domain adaptation that aims to minimize the difference between the source and target domains \cite{gretton_covariate_2009}. Domain adaptation has been shown to improve the generalisability of classifiers for the same task across different domains  \cite{tzeng_simultaneous_2015, tzeng_adversarial_2017}. We draw inspiration from domain adaptation methods to make the network agnostic to spurious variations.

Learning feature representations that are invariant to certain spurious variations has been tackled in a number of ways.  \cite{yin_multi-task_2018} take a multi-task learning approach to pose-invariant face recognition. They propose a novel approach to automatically weight each of the spurious variations during training. 
Another approach is to adjust the training data distribution at training time, to avoid learning biases \cite{zhao_men_2017}. This method relies on having labels for each of the spurious variations for each training instance. This is not feasible for most existing datasets, as they tend to be labeled with a single task in mind, where information about spurious variations is not available. Our method can make use of separate datasets, each labeled for distinct tasks, to remove multiple spurious variations simultaneously.

Jia et al \cite{jia_right_2018} and Raff et al \cite{raff_gradient_2018} draw inspiration from \cite{ganin_domain-adversarial_2015} to remove a source of variation with the use of a gradient reversal layer to update the network in opposition of a task. Instead of applying gradient reversal on the output of the softmax layer, which penalizes correct classifications, we compute the cross-entropy of the output classifier and a uniform distribution, as in \cite{tzeng_simultaneous_2015, tzeng_adversarial_2017}. This ensures an equally uninformative classifier across all tasks.

% ------------------------------------------------------------
\section{\label{sec:data}Datasets}
% ------------------------------------------------------------

\subsection{Labeled Ancestral Origin Faces in the Wild}
\label{subsec:LAOFIW}

A new ancestral origin database was created as part of this work called ``Labeled Ancestral Origin Faces in the Wild (LAOFIW)''. The aim of this dataset was to 1) create a dataset for experimentation, and 2) be used as a spurious variation dataset for applications where training a network to be agnostic to ancestral origin is important.
The database was assembled using the Bing Image Search API. Search terms based upon origin, e.g. ``German, English, Polish, etc'', were submitted in conjunction with the words ``man, woman; boy, girl''. Additionally, results were filtered to return photographic images of faces. In total, 43 origin search terms were queried returning 20,000 images. Duplicates were removed by comparing their Histogram of Oriented Gradients \cite{dalal_histograms_2005} encoded using a Gaussian mixture model for each image. 
The remaining 14,000 images, were manually divided into four broad ancestral origins: sub-Saharan Africa, Indian Subcontinent, Europe, and East Asia. These classes were selected on the basis of being visually distinct from each other. 

\begin{figure}[h]
  \begin{center}
  \includegraphics[width=0.8\textwidth]{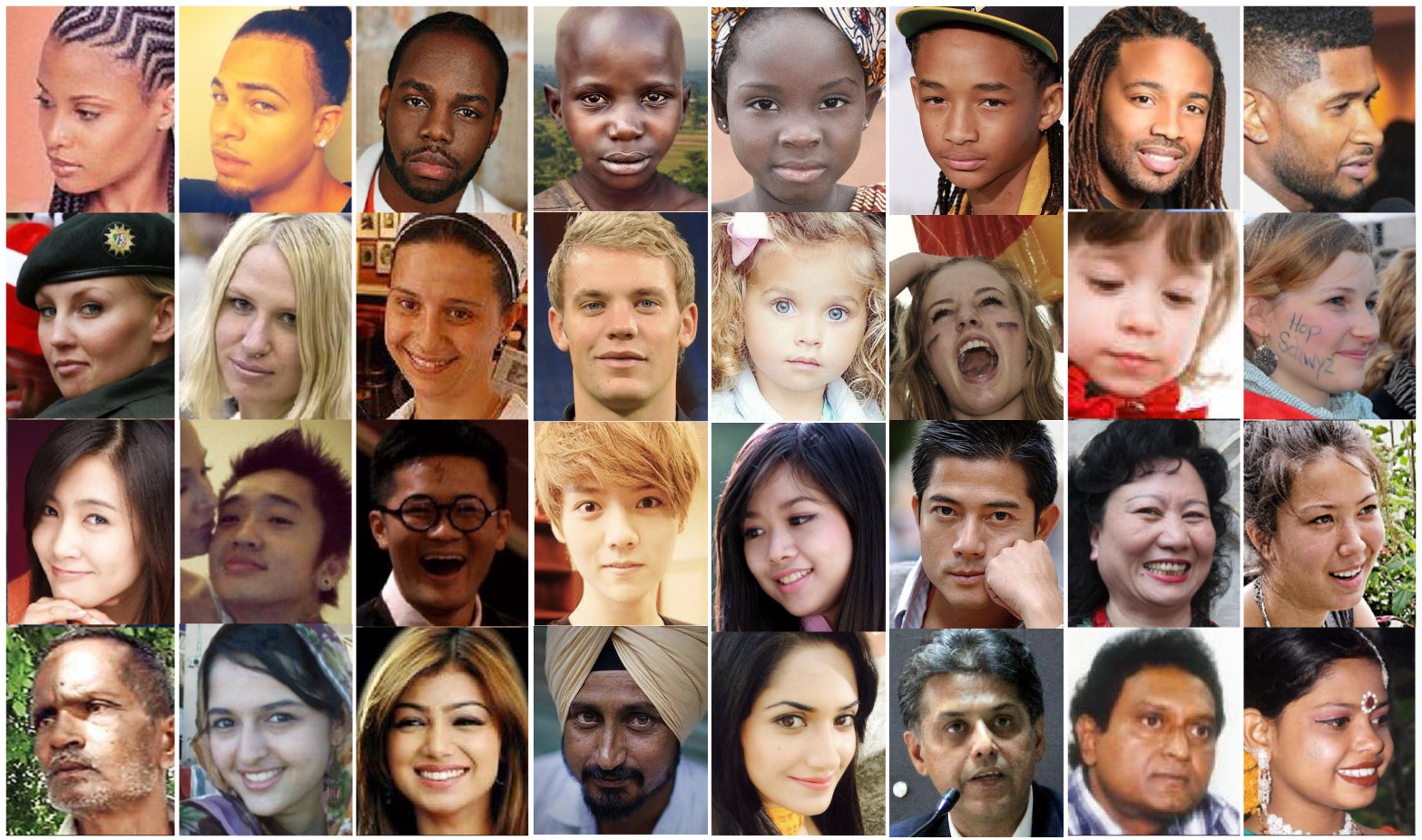}
  \caption{Example images from the LAOFIW dataset for each of the four classes. In rows top to bottom: Sub-Saharan Africa, Europe, East Asia, Indian Subcontinent. The images are highly varied in age, gender, pose, lighting and expression.}
  \label{fig:LAOFIWexamples}
  \end{center}
  \vspace{-0.5cm}
\end{figure}

The database contains roughly the same number of male and female individuals. It also contains images with varied poses: more than a third of the images are non-frontal. Sample images are shown in figure \ref{fig:LAOFIWexamples}.

\subsection{Age and Gender Dataset}
\label{subsec:imdb}

The IMDB dataset is a large publicly available dataset of the images contained in the profiles of the 100,000 most popular actors on the IMDb website\footnote{\url{https://www.imdb.com/}} with gender and date of birth labels provided for each instance \cite{rothe_dex_2015}. We used this dataset to investigate the effects of age and gender bias in celebrity datasets. 

The labels in the IMDB dataset are noisy, with a number of individuals having both incorrect age and gender labels. This is due to the nature of how the data was collected: The authors assumed that images from an actor's profile that contained a single face would show the actor in question. As stated by the authors, however, these images often contain other actors, who co-starred in their movies \cite{rothe_dex_2015}. As gender and date of birth are taken from the profile of the actor, this causes erroneous labels for images that show co-stars. The age of an individual was calculated as the difference between the timestamp of the photo and the date of birth of the subject. In some cases, the time stamps of photos predate an actor's date of birth or are otherwise unreliable.

To mitigate this problem, we used the Microsoft Azure Face API\footnote{\url{https://azure.microsoft.com/en-gb/services/cognitive-services/face/}} to extract gender and age estimates for an identity-balanced subset of 150,000 images. We rejected all images in which the predicted gender from Azure and the IMDB gender label disagreed. We then ran the analysis for age and removed images in which the Azure age prediction differed by more than 10 years from the IMDB labels. The resulting, cleaned dataset contained 60,000 images.

In order to quantify the effectiveness of this data cleansing procedure, we trained a gender classification VGG-M network \cite{simonyan_very_2014} on both the original 150,000 images and the cleaned 60,000 images, withholding 20\% of the images for testing. The gender classification accuracies on the test images improved from 75\%, before cleaning, to 99\% after cleaning.

\begin{figure}[h]
  \begin{center}
  \includegraphics[width=0.5\textwidth]{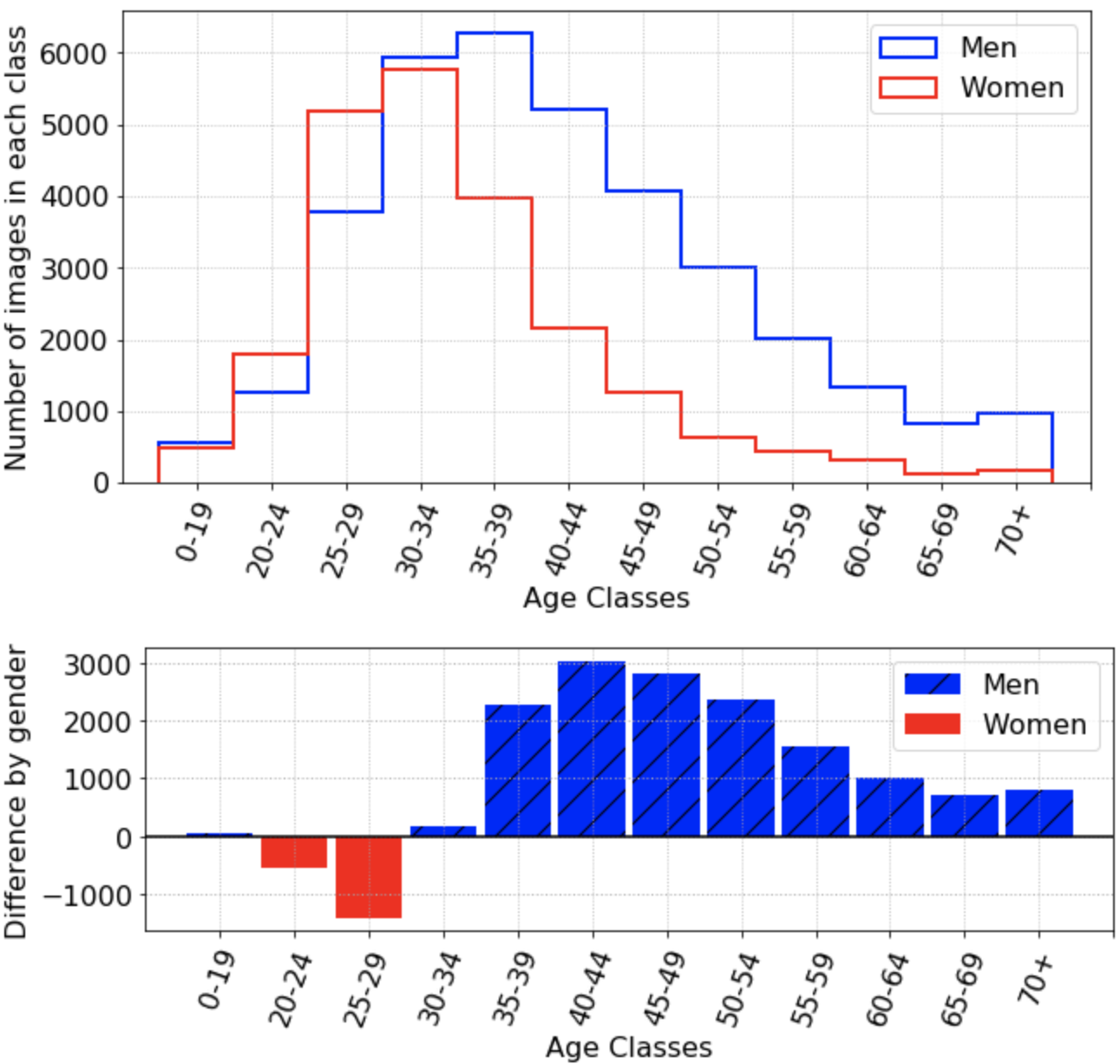}
  \caption{Age distributions and differences for women (red) and men (blue) in the IMDB dataset. The dataset bias towards younger women and older men is visible.}
  \label{fig:imdbhist}
  \end{center}
  \vspace{-1cm}
\end{figure}

The distribution of ages for each gender in the cleaned IMDB dataset is shown in figure \ref{fig:imdbhist}. We can observe a bias towards younger women and older men in the data.

A subset of the cleaned data was used to create an unbiased test dataset, which contains equal numbers of men and women for each age category. This unbiased test dataset was used to evaluate network bias and was not used during training.

\subsection{Pose Dataset}
\label{subsec:aflw}

We used the AFLW dataset \cite{kostinger_annotated_2011} categorized by yaw, to train the pose classifier mentioned in section \ref{subsec:LAOFIW}. The poses were binned into five categories, ``profile left, near-frontal left, frontal, near-frontal right, profile right''. The non-frontal images were duplicated and flipped to augment the dataset. A class-balanced subset of 24,000 images was selected for training, and a class-balanced set of 6,000 images was reserved for testing.

% ------------------------------------------------------------
\section{\label{sec:methods}Methods}
% ------------------------------------------------------------
We introduce a supervised-learning algorithm that aims to learn a single feature representation $\theta_{\text{repr}}$, that is informative for a primary classification task, whilst simultaneously being uninformative for a number of spurious variations, that represent undesirable sources of variation. For example, we may wish to create an age classifier, that does not base its decisions on any pose, ancestral origin, or gender information.

We assume access to a primary dataset $\mathcal{D}_p = \left\{x_i, y_i\right\}_{i=1}^{n_p}$, containing $n_p$ images, $x_i$, with labels $y_i \in \{1, ..., K\}$. And similarly, we assume access to $M$ secondary datasets, $\mathcal{D}_s = \{\mathcal{D}_m\}_{m=1}^M$, each describing a single spurious variation.

\subsection{Joint Learning and Unlearning}
\label{subsec:learning}

We introduce a joint learning and unlearning (JLU) algorithm to learn a primary classification task, whilst simultaneously unlearning multiple spurious variations.

\begin{figure}[h]
        \begin{center}
        \includegraphics[width=0.8\textwidth]{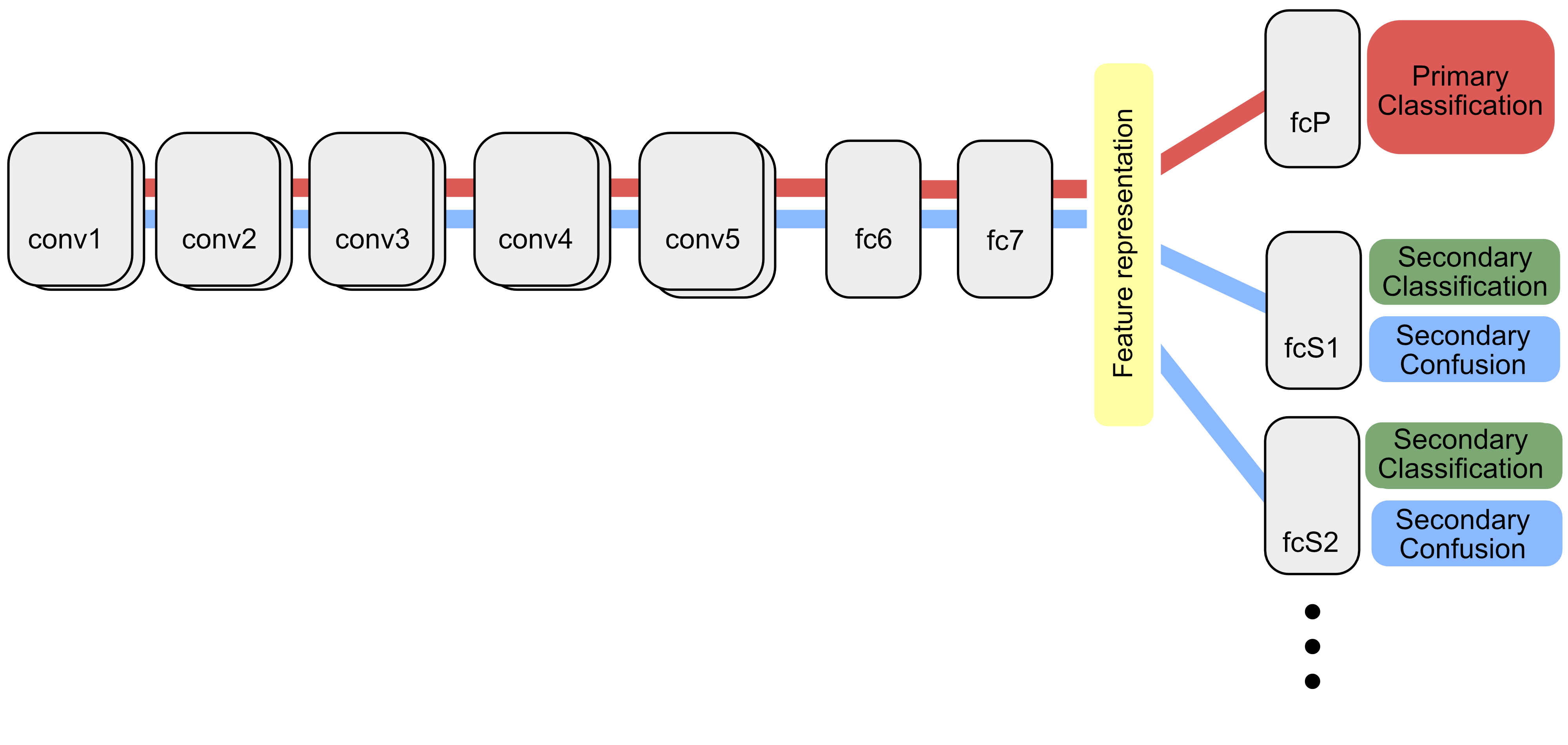}
        \caption{Overview of CNN Architecture: We use a joint loss over primary and secondary data to learn a feature representation that simultaneously learns to classify the primary task but becomes invariant to secondary tasks, the spurious variations. The spurious variation classification loss and confusion loss act in opposition to learn the classifier on the feature embedding and change the feature embedding to confuse the classifier, respectively. The base architecture is a VGG-M network \cite{simonyan_very_2014}.}
        \label{fig:net}
        \end{center}
        \vspace{-0.3cm}
\end{figure}

Our convolutional network (CNN) architecture is depicted in Figure \ref{fig:net}. The primary branch has a single classification loss, the primary classification loss, which assesses the ability of the network to accurately distinguish between classes of the primary task. Each secondary branch has two losses: a classification loss and a confusion loss. These losses are used to, in turn, assess the amount of spurious variation information left in the feature representation $\theta_\text{repr}$ and then remove it.

Let the classification objective for a generic task with $K$ classes and a corresponding classifier $\theta_c$, given a feature representation $\theta_{\text{repr}}$, be defined as the standard softmax loss:

\begin{equation}
L_{\text{softmax}}(x, y; \theta_{\text{repr}}, \theta_{C}) = - \sum_{k=1}^K \mathds{1} [y = k] \log p_k,
\label{eq:class}
\end{equation}

where $p_k$ is the softmax of the classifier's output assigning the input $x$ to class $k$. We will refer to this loss evaluated on the primary task as $L_{P}(x_p, y_p; \theta_{\text{repr}}, \theta_{P})$ and on the $m$-th spurious variation as $L_{m}(x_{m}, y_{m}, \theta_{\text{repr}}; \theta_{m})$.

Inspired by \cite{tzeng_simultaneous_2015}, we introduce a confusion loss for the $m$-th spurious variation, $L_{\text{conf}}(x_m, y_m, \theta_m; \theta_{\text{repr}})$, in (\ref{eq:conf}). Minimizing the confusion loss seeks to change the feature representation $\theta_{\text{repr}}$, such that it becomes invariant to the spurious variations.
\begin{equation}
L_{\text{conf,m}}(x_m, y_m, \theta_{m}; \theta_{\text{repr}}) = - \sum_{n_m} \frac{1}{n_m} \log p_{n_m} .
\label{eq:conf}
\end{equation}
We compute the best classifier for each spurious variation, $\theta_{m}$, and then compute the entropy between the output predicted from each of these classifiers and a uniform distribution. The complete method minimizes the joint loss function:
\begin{equation}
\begin{aligned}
L(x_p,y_p,x_s,y_s,\theta_p, \theta_s, \theta_{\text{repr}}) &= 
 L_{p}(x_p, y_p; \theta_{\text{repr}}, \theta_{p}) \\
&+ L_{s} 
 + \alpha L_{\text{conf}},
\label{eq:joint}
\end{aligned}
\end{equation}
where $\alpha$ determines how strongly the confusion loss affects the overall loss, $\theta_s = \{\theta_1,...,\theta_M\}$, and
\begin{align}
    L_{\text{conf}} &= \frac{1}{M} \sum_{m=1}^M L_{\text{conf}, m}(x_m, y_m, \theta_{m}; \theta_{\text{repr}})
    \label{eq:secondary_conf_loss} \\
    L_s &= \sum_{m=1}^M \beta_m L_m(x_m, y_m, \theta_{\text{repr}}; \theta_{m}),
    \label{eq:secondary_classif_loss}
\end{align}
where $\beta_m$ is a weight assigned to the $m-$th spurious variation.
As mentioned in \cite{tzeng_simultaneous_2015}, the confusion loss (\ref{eq:secondary_conf_loss}) and the spurious variation classification loss (\ref{eq:secondary_classif_loss}) stand in opposition to one another, so they cannot be optimized in the same step. Therefore, we switch between training the spurious variation classification loss, $L_{s}$, and then the joint primary and confusion loss, $L_{p}(x_p, y_p; \theta_{\text{repr}}, \theta_{p}) + \alpha L_{conf}$. At each iteration, we find the best spurious variation classifier for the feature representation. The training procedure is shown in algorithm \ref{alg:joint}.
We used the VGG-M architecture \cite{chatfield_return_2014}, which consists of five convolutional layers (conv1-5) and three fully connected layers (fc6-8). The feature representation parameters $\theta_{\text{repr}}$ represents layers conv1-fc7.

\begin{algorithm}[h]
        \caption{Joint Learning and Unlearning}\label{euclid}
        \begin{algorithmic}[1]
          \Procedure{$\min$ L}{$x_p,y_p,x_s,y_s,\theta_p, \theta_s, \theta_{\text{repr}}$}
          \For{epochs}
            \While{$\frac{dL_s}{d\theta_s} \not= 0$}
              \State $\min L_s$
            \EndWhile
            \State $\min L_p + \alpha L_\text{conf}$
         \EndFor
          \EndProcedure
        \end{algorithmic}
        \label{alg:joint}
\end{algorithm}
\vspace{-0.5cm}

% ------------------------------------------------------------
\section{\label{sec:experiments}Experiments}
% ------------------------------------------------------------
In this section, we present experiments to demonstrate possible applications of our methodology and the datasets we tested them on.

\begin{description}
        \item [Removal of a bias from a network ---] We train a gender-agnostic age classifier using the IMDB dataset, which contains a gender bias: female celebrities tend to be younger than their male counterparts in this dataset.
        \item [Removal of an extreme bias from a network ---] We train an age-agnostic gender classifier on subsets of the IMDB dataset that contain only young women and only old men, and vice versa.
        \item [Simultaneous removal of multiple spurious variations ---] We demonstrate our JLU algorithm's ability to simultaneously remove multiple spurious variations from the feature representation of a network trained for a primary classification task.
\end{description}

\subsection{Removal of a bias from a network}
\label{subsec:expremovebias}
We investigated the task of creating an age classifier using the cleaned IMDB dataset described in section \ref{subsec:imdb}.
%  (the age classification approach is detailed in section \ref{subsec:age})
We hypothesize that the gender bias in the distribution, where men are generally older than their female counterparts in this dataset, will be learned by the network. 
% The network should learn to underestimate the ages of women and overestimate the ages of men.

We trained two networks to perform this task: 1) baseline --- trained solely on age data, 2) blind --- trained on age data, whilst removing gender-specific information from the network.

We evaluated both networks on the unbiased test dataset detailed in section \ref{subsec:imdb}. We compared accuracies for age classification and prediction distributions for both genders.

\subsection{Removal of an extreme bias from a network}
\label{subsec:expremovepbias}

\begin{figure}[h]
    \centering
    \begin{subfigure}[b]{0.48\textwidth}
        \includegraphics[width=\textwidth]{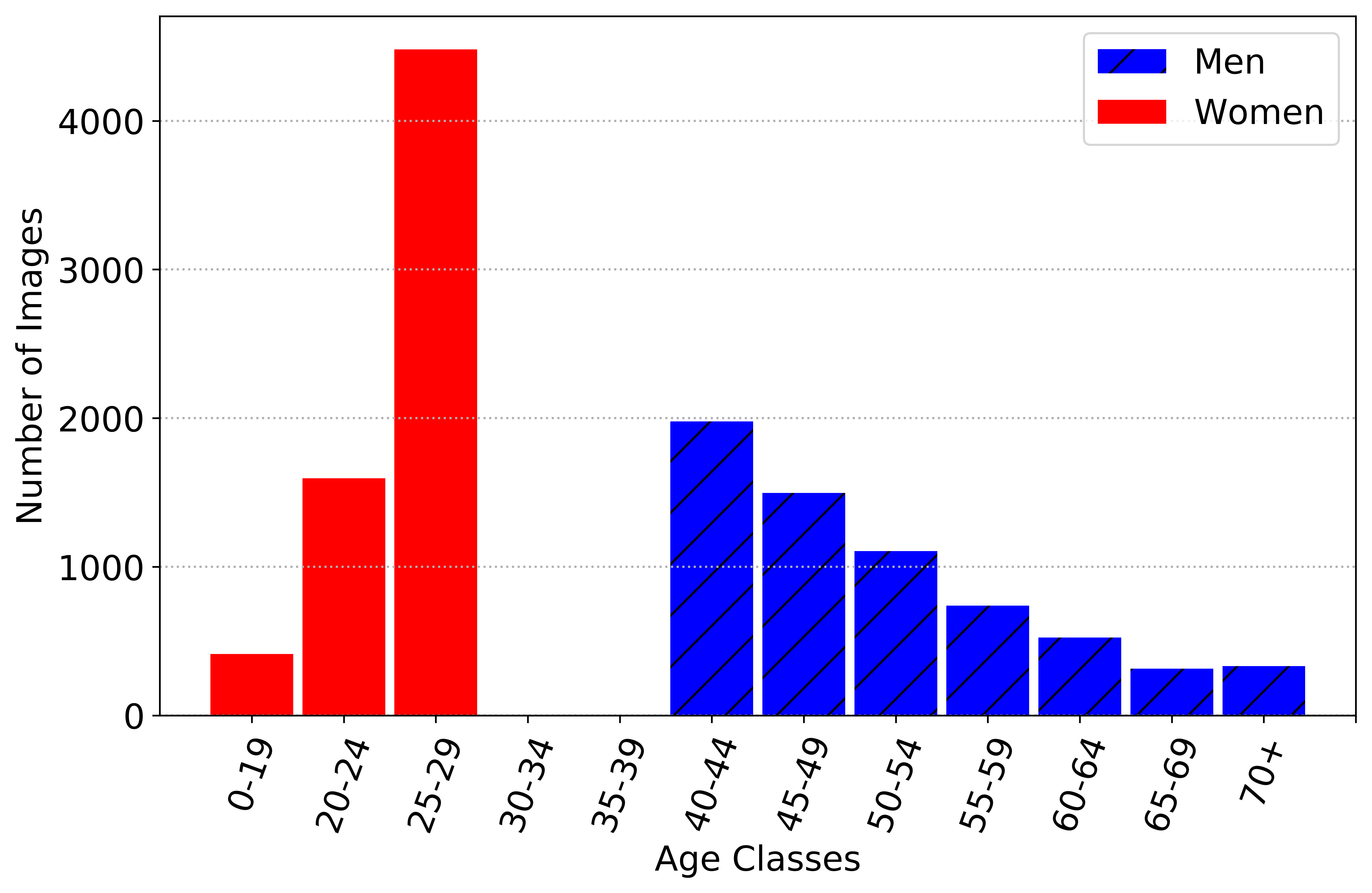}
        \caption{EB1: women aged 0-30, men aged 40+}
    \end{subfigure}
    ~ %add desired spacing between images, e. g. ~, \quad, \qquad, \hfill etc. 
    \begin{subfigure}[b]{0.48\textwidth}
        \includegraphics[width=\textwidth]{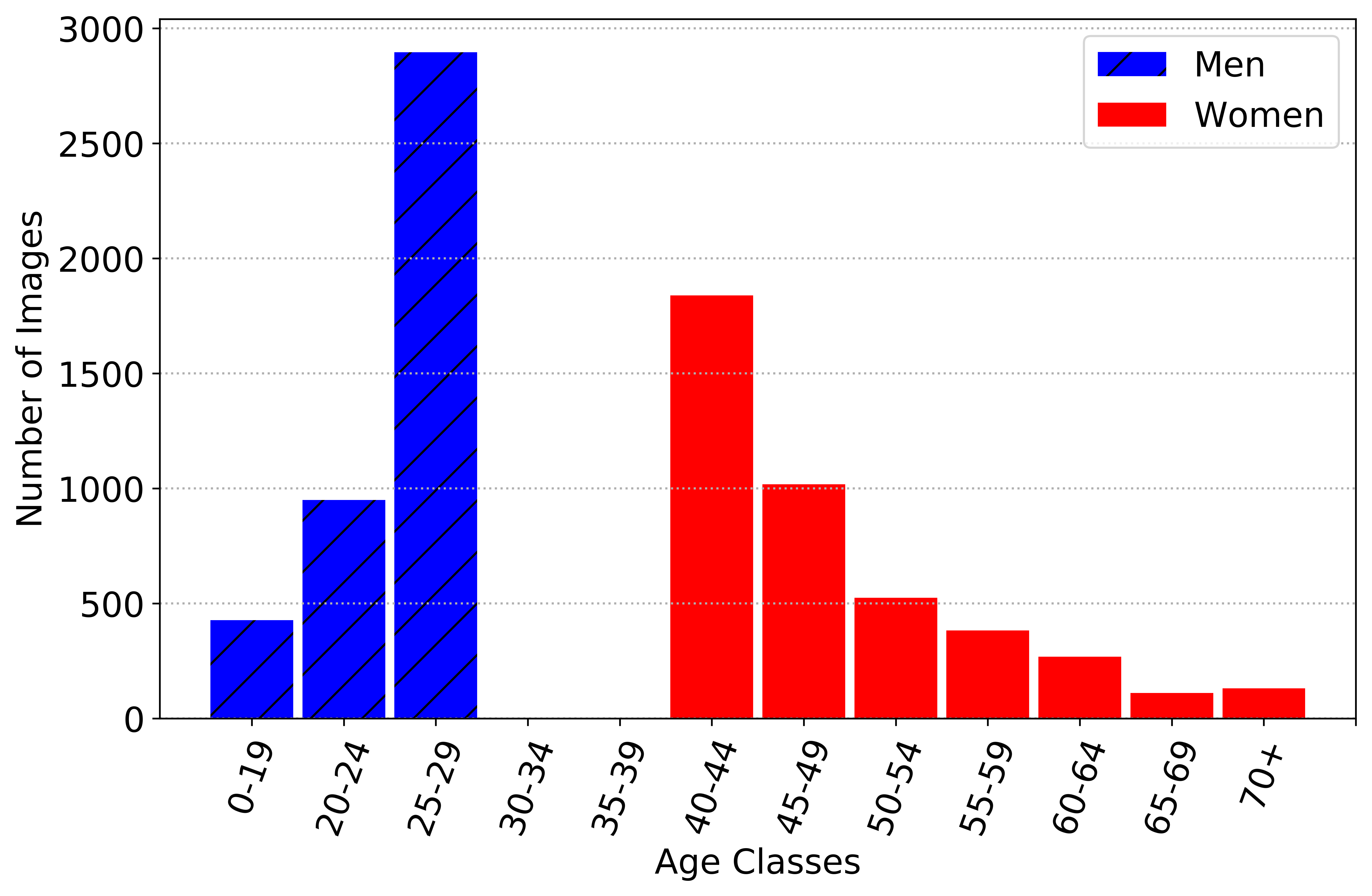}
        \caption{EB2: women aged 40+, men aged 0-30}
    \end{subfigure}
    \caption{Histograms showing the ages of women (red) and men (blue) in the artificially biased datasets. In both cases, individuals aged under 35 were selected for the young, and individuals aged above 45 were selected for the old set.}
    \label{fig:poison}
\end{figure}

We train gender classifiers using the following artificially biased subsets of the cleaned IMDB data from section \ref{subsec:imdb}:
\begin{enumerate}
    \item Extreme bias 1 (EB1): women aged 0-30, men aged 40+
    \item Extreme bias 1 (EB2): women aged 40+, men aged 0-30
\end{enumerate}
We placed a buffer of 10 years without any individuals between the two sets to exaggerate the age difference between the genders. We hypothesize that the age bias in the distribution will be learned by the network, for example, if trained on EB1, it should falsely learn to predict that young men are in fact women. Histograms of these datasets are given in figure \ref{fig:poison}.

We trained a baseline network on gender only, and a blind network using our JLU algorithm to unlearn age.
The trained classifiers were evaluated on the unbiased test dataset detailed in section \ref{subsec:imdb}.

\subsection{Simultaneous removal of multiple spurious variations}
\label{subsec:expremovemulti}
We demonstrate our algorithm's ability to remove multiple spurious variations from the feature representation of a network. The primary task and spurious variations are selected from age, gender, ancestral origin and pose. In each case, one task is selected as the primary task, and the others make up the spurious variations that we wish to remove. A baseline network was trained for each primary task without using our JLU algorithm.

We used the cleaned IMDB dataset described in section \ref{subsec:imdb} for age and gender labels. We used our LAOFIW dataset, described in section \ref{subsec:LAOFIW} to classify ancestral origin. Finally, we used the adapted AFLW dataset (Section \ref{subsec:aflw}) to classify pose.

\subsection{\label{subsec:age}Age Classification}
Since our dataset is relatively small for a deep learning task, we do not approach age estimation as a regression task. We conduct age classification by creating age bins of 5 years, or more years at either end of the distribution. The categories are shown in figure \ref{fig:imdbhist}. We define predictions within one class from the true age class as a positive classification to account for errors caused by edge cases in different bins.

\subsection{Implementation details}
Our base network is an adapted VGG-M network \cite{simonyan_very_2014}, pre-trained on the VGG-Face dataset \cite{parkhi_deep_2015}. In our experiments, we saw no significant improvement in updating the weights in the convolutional layers, therefore, the weights in layers conv1-conv5 were frozen for all experiments. This significantly increased the speed of the training algorithm.

The confusion loss was approximated using the Kullback-Leibler divergence between the softmax output of a classifier and a uniform distribution. The network is trained using stochastic gradient descent (SGD) with a learning rate of $1\times10^{-4}$. Learning rates of classification layers for each task were boosted by a factor of 10. The hyperparameter in equation \ref{eq:joint} is set to a value of $\alpha=0.1$. The spurious variation classifier hyperparameters were all set to $\beta_m=1$. To address the imbalance in the class distributions of the training data, losses were weighted by the inverse of the relative frequency of that class. We conduct our experiments using the MatConvNet framework \cite{vedaldi15matconvnet}.

% ------------------------------------------------------------
\section{\label{sec:results}Results}
% ------------------------------------------------------------

\subsection{Removal of a bias from a dataset}

\begin{figure}[h]
    \centering
    \begin{subfigure}[b]{0.4\textwidth}
        \includegraphics[width=\textwidth]{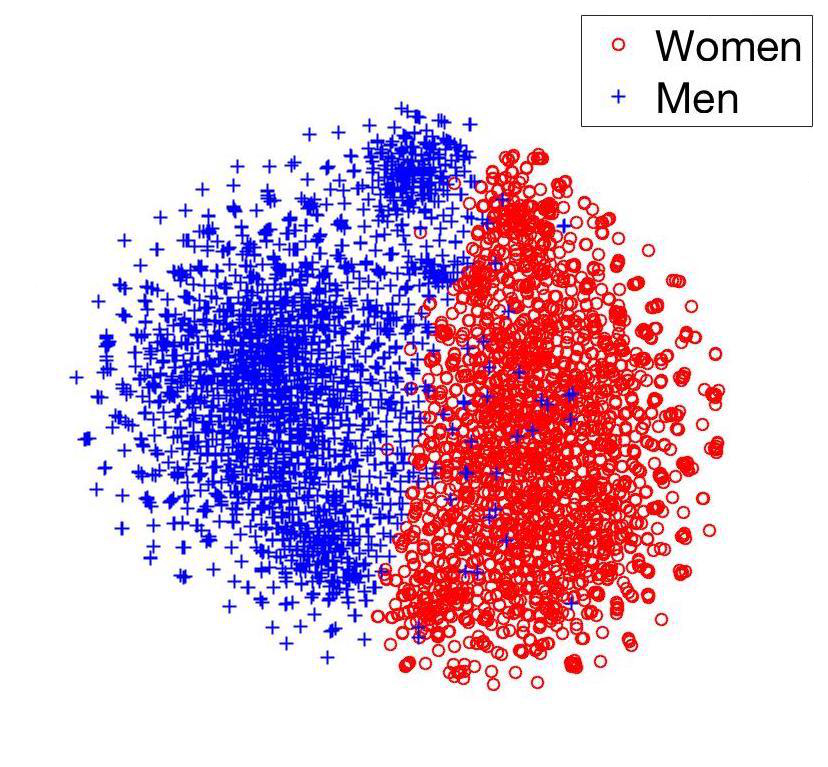}
        \caption{Feature representation of baseline network}
        \label{fig:agebase}
    \end{subfigure}
    ~\qquad 
    \begin{subfigure}[b]{0.4\textwidth}
        \includegraphics[width=\textwidth]{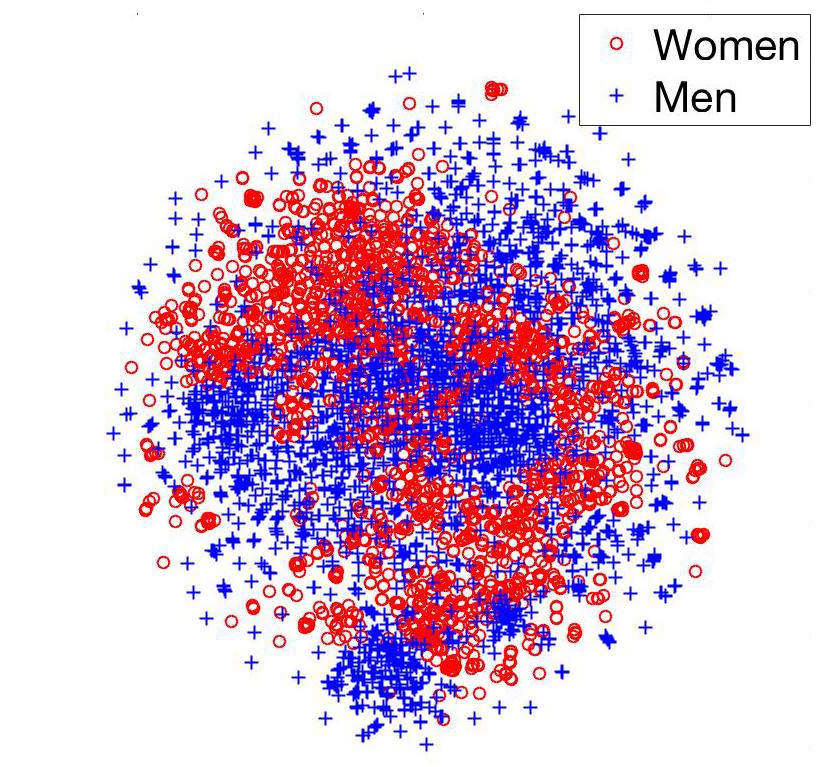}
        \caption{Feature representation of blind network}
        \label{fig:ageunl}
    \end{subfigure}
    \caption{T-SNE visualizations of 4096-dimensional feature representation of the class-balanced test dataset for age classification networks trained on gender-biased data. (a) Baseline network without unlearning gender, (b) Blind network with gender unlearning. The feature representation is clearly separable by gender for the baseline network, showing that the network has learned the gender bias in the dataset. After unlearning, this bias is no longer pronounced.}
    \label{fig:tsne}
\end{figure}

We computed feature representations for the class-balanced test dataset from section~\ref{subsec:imdb} for both baseline and blind networks. T-SNE visualizations of these feature embeddings are shown in figure \ref{fig:tsne}. The feature representation of the baseline network that was trained to classify age is clearly separable by gender, demonstrating that the bias in the training data was learned. After unlearning gender, the feature representation is no longer separable by gender, demonstrating that this bias has been removed.

\begin{figure}[h]
    \centering
    \begin{subfigure}[b]{0.45\textwidth}
        \includegraphics[width=\textwidth]{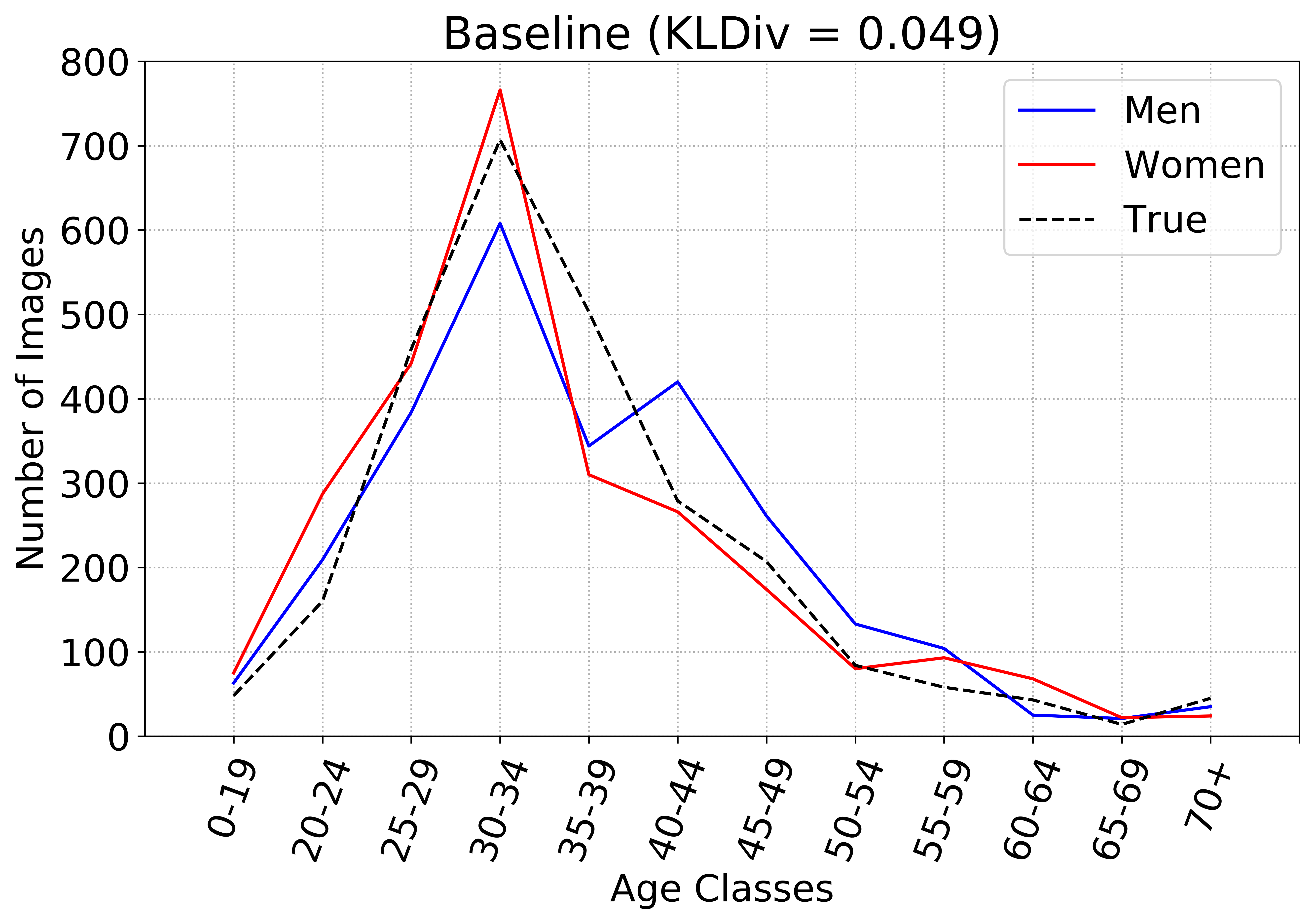}
        \caption{Baseline network}
        \label{fig:agebase}
    \end{subfigure}
    ~ %add desired spacing between images, e. g. ~, \quad, \qquad, \hfill etc. 
    \begin{subfigure}[b]{0.45\textwidth}
        \includegraphics[width=\textwidth]{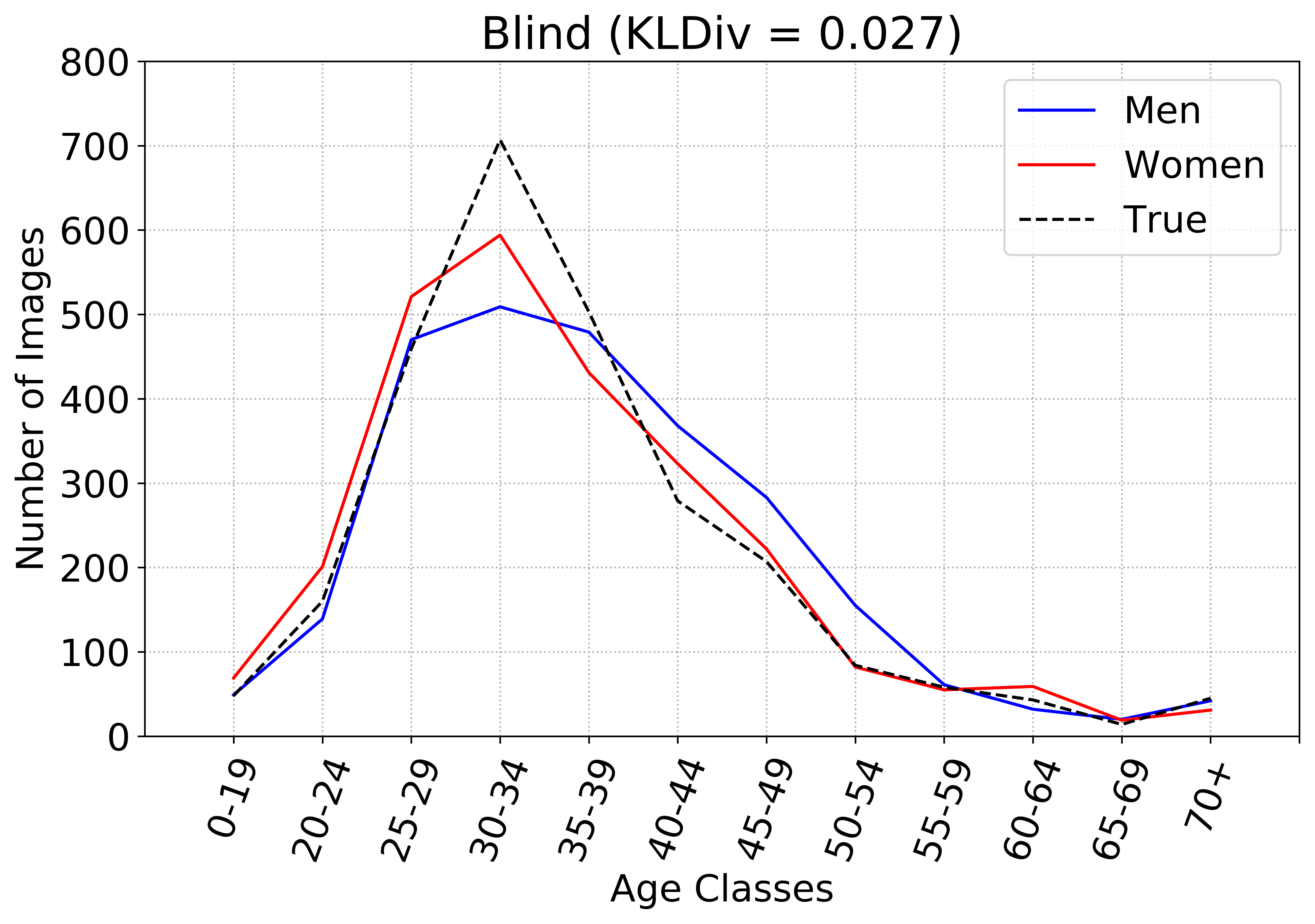}
        \caption{Unbiased network}
        \label{fig:ageunl}
    \end{subfigure}
    \caption{Prediction distributions of age for each gender compared to the ground truth (black) for men (blue) and women (red). The evaluation dataset was gender-balanced so that each class has an equal number of women and men. (a) Baseline network without unlearning gender, (b) unbiased network with gender unlearning. The red line shows the age predictions for women and the blue line shows them for men. The prediction distributions for women and men align after unlearning gender. The KL-divergence score between the prediction distributions of men and women are given for each case. The KL-Divergence score reduces, as two distributions become more similar.}
    \label{fig:singlesov}
    % \vspace{-1cm}
\end{figure}

The distributions of the age predictions and ground truth on the gender-balanced test set are shown in figure \ref{fig:singlesov}. The red and blue lines show the predicted ages for women and men, respectively. The black dashed line shows the ground truth for both genders. The distributions of predicted ages for females and males are different for the baseline network (Figure \ref{fig:agebase}): women's ages tend to be underestimated, whereas men's ages are generally overestimated. This behaviour mimics the bias within the training data, which confirms that the classification model is making age predictions that are, to a degree, dependent on gender. 

In order for us to be sure that the network is not using gender information to predict age, both female and male prediction distributions should be similar. Figure \ref{fig:ageunl} shows the distributions for the network that has been trained to no longer be able to differentiate between genders. The KL-divergence between age prediction distributions for men and women of the biased and unbiased networks are 0.049 and 0.027, respectively. The reduction in KL-divergence shows that the network has successfully unlearned gender, as the predicted age distributions for both genders are similar.
Note, that we are not necessarily trying to perfectly predict the ground truth, but be confident that we are not treating men and women differently.

The average prediction accuracy for age classification on the gender-balanced test dataset for the baseline network was 78.9\% (78.4\% for females and 79.4\% for males) and for the unbiased network was 78.1\% (77.4\% for females and 78.9\% for males). This corresponds to a reduction in accuracy of 0.8\% (1.2\% for females and 0.5\% for males).

% ------------------------------------------------------------
\subsection{Removal of an extreme bias from a network}

The baseline gender classification accuracy on the gender-balanced test data was 70\% for a network trained on the EB1 dataset. This was improved by 16\% to 86\% for the blind network, where we simultaneously unlearned age information. The accuracies for networks trained on the EB2 dataset were 62\% for the baseline and 82\% for the blind network. This amounts to a 20\% increase in classification accuracy.

\begin{figure}[h]
    \centering
    \begin{subfigure}[b]{0.9\textwidth}
        \includegraphics[width=\textwidth]{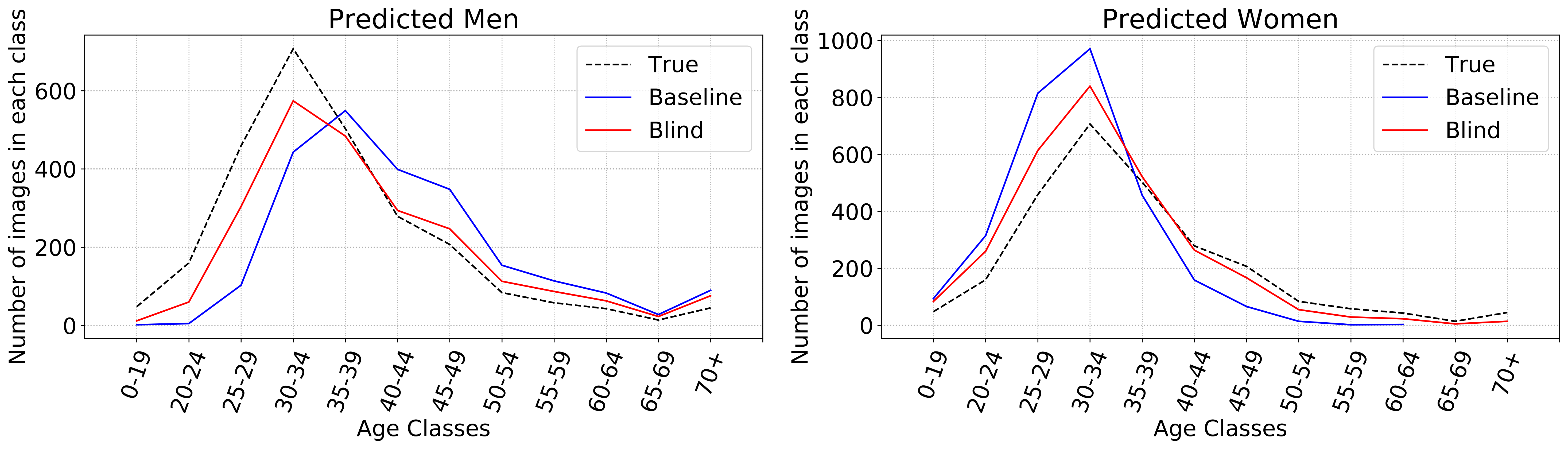}
        \caption{Male and female predictions by age by network trained on EB1 dataset}
        \label{fig:extremebiasresult1}
    \end{subfigure}
    \par\bigskip
    \begin{subfigure}[b]{0.9\textwidth}
        \includegraphics[width=\textwidth]{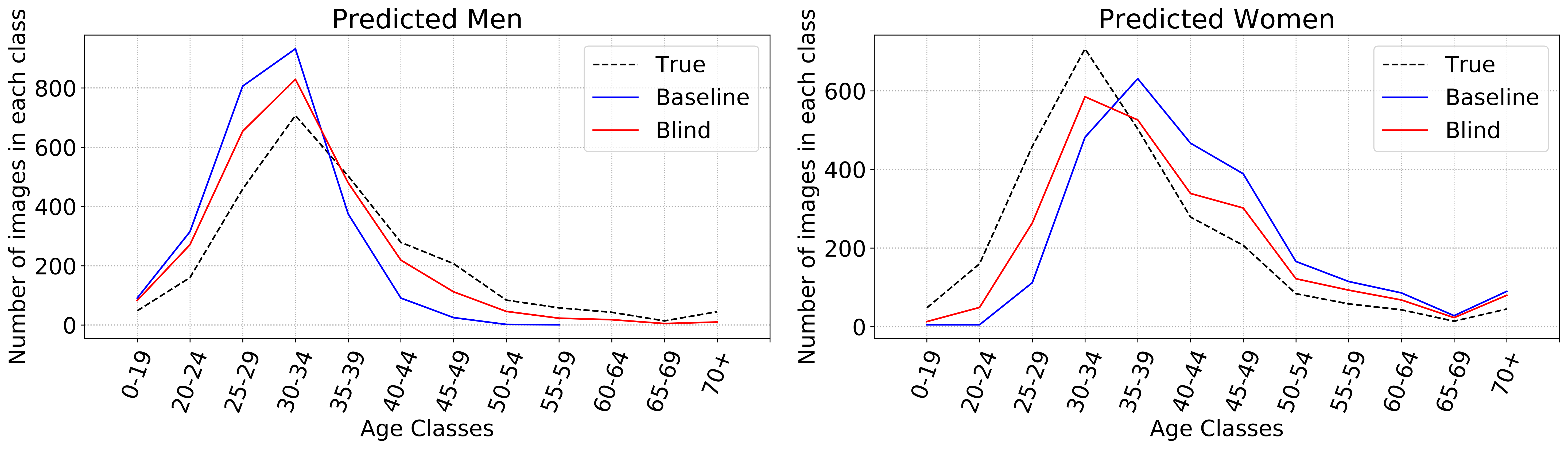}
        \caption{Male and female predictions by age by network trained on EB2 dataset}
        \label{fig:extremebiasresult2}
    \end{subfigure}
    \caption{Age distributions of male and female predictions from networks trained on the biased (a) EB1 and (b) EB2 datasets, evaluated on the gender-balanced test data.
    The baseline network often wrongly predicts older individuals to be male and younger individuals to be female. Using the JLU algorithm reduces this bias and shifts the prediction distributions closer to the true distribution.}
    \label{fig:extremebiasresult}
    % \vspace{-1cm}
\end{figure}

The age distributions of individuals that were predicted to be either male or female, trained on the EB1 dataset, are shown in figure \ref{fig:extremebiasresult}. The baseline network often wrongly predicts older individuals to be male, and younger individuals to be female, in line with the training data. The age distribution of gender predictions from the JLU network, however, is closer to the true distribution. We observed similar results for the baseline network trained on the EB2 dataset, where younger individuals were more often predicted to be male, and older individuals were more often predicted to be female. The JLU network prediction distributions were closer to the true distributions.

% ------------------------------------------------------------
\subsection{Simultaneous removal of multiple spurious variations}

\begin{figure}[h]
    \centering
    \begin{subfigure}[b]{0.47\textwidth}
        \includegraphics[width=\textwidth]{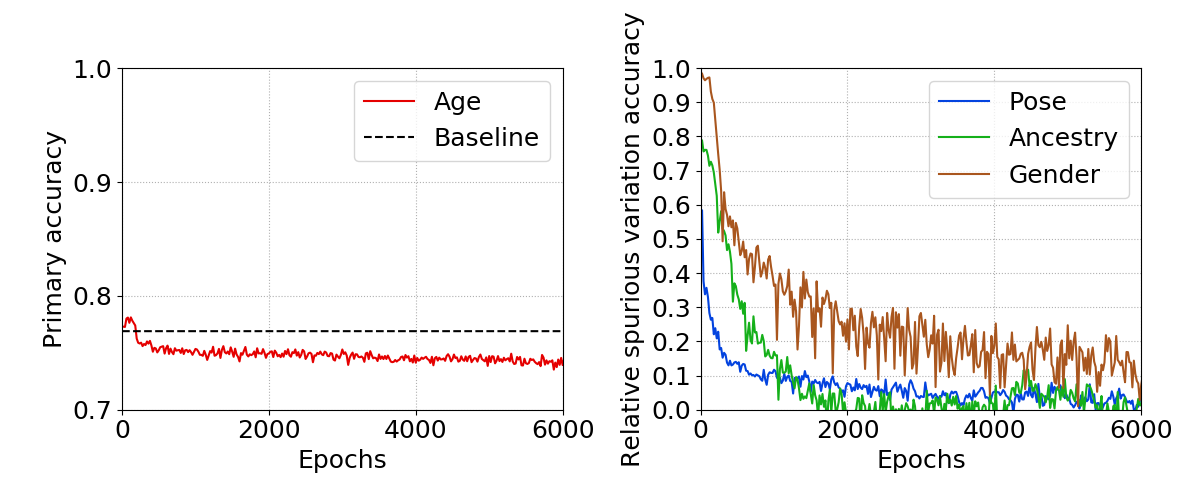}
        \caption{\textbf{Age} --- Ancestral Origin, Gender, Pose}
        \label{fig:multi_age}
    \end{subfigure}
    ~
    \begin{subfigure}[b]{0.47\textwidth}
        \includegraphics[width=\textwidth]{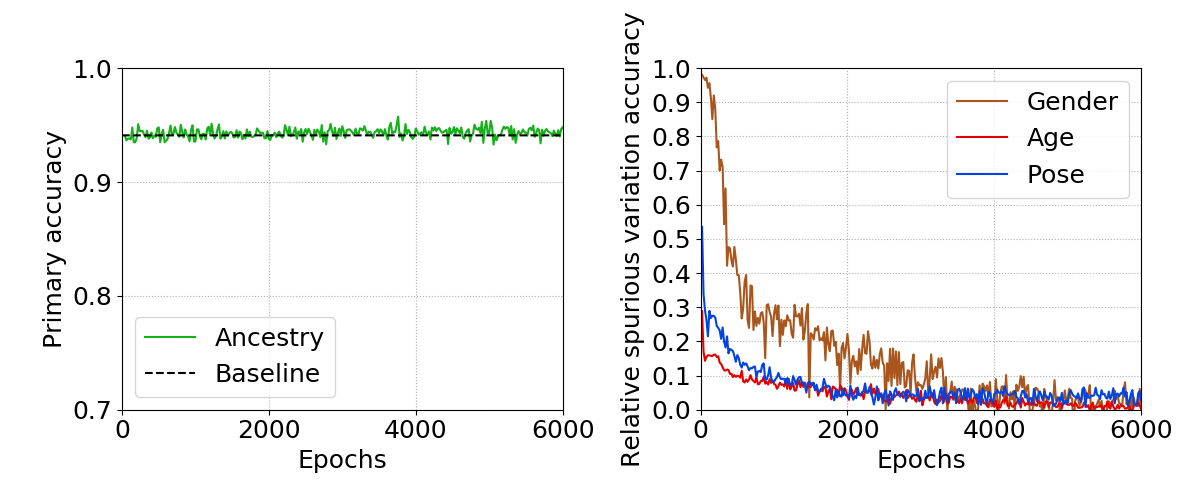}
        \caption{\textbf{Ancestral Origin} --- Age, Gender, Pose}
        \label{fig:multi_anc}
    \end{subfigure}
    \par
    \begin{subfigure}[b]{0.47\textwidth}
        \includegraphics[width=\textwidth]{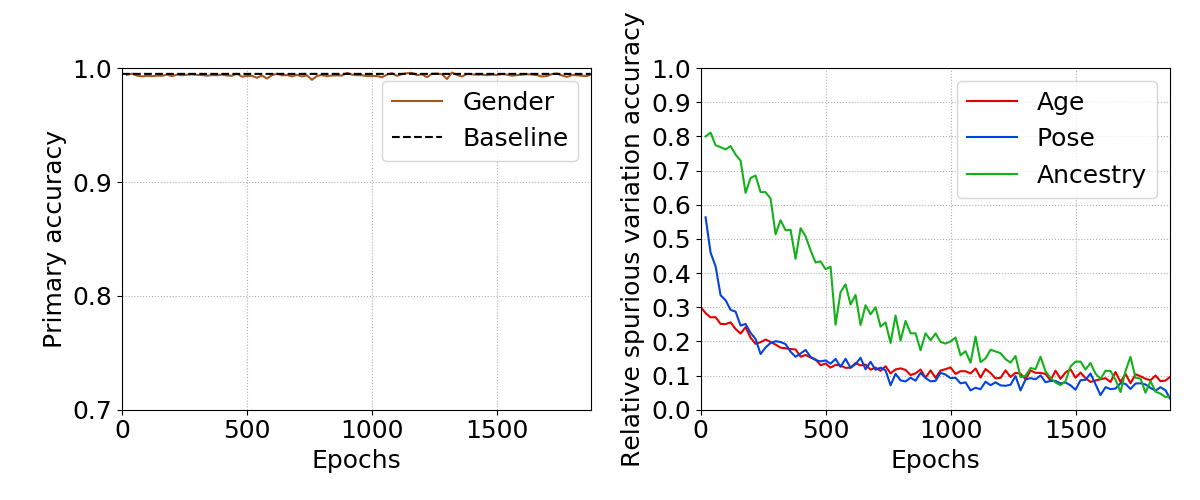}
        \caption{\textbf{Gender} --- Age, Ancestral Origin, Pose}
        \label{fig:multi_gender}
    \end{subfigure}
    ~
    \begin{subfigure}[b]{0.47\textwidth}
        \includegraphics[width=\textwidth]{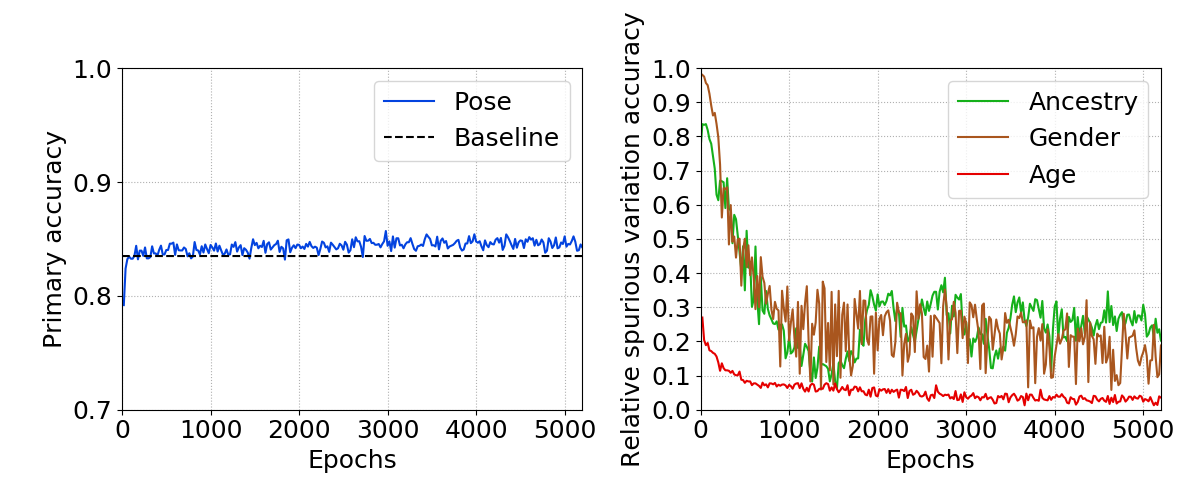}
        \caption{\textbf{Pose} --- Age, Ancestral Origin, Gender}
        \label{fig:multi_pose}
    \end{subfigure}
    \caption{Classification accuracies on test data for the primary task and spurious variations during the training procedure of the JLU algorithm. The primary classification accuracy is compared to the baseline network accuracy (dotted). The spurious variation accuracies have been re-scaled using equation \ref{eq:rescale} so that zero accuracy corresponds to the accuracy of a random chance classifier.}
    \label{fig:multi}
    % \vspace{-1cm}
\end{figure}

Figure \ref{fig:multi} shows how the accuracies on the test data for each attribute vary over the JLU training procedure when simultaneoulsy unlearning multiple spurious variations. For clarity, in the figure, spurious variation classification accuracies were rescaled using the equation below:
\begin{equation}
    a = 1 - \frac{e}{e_{\text{max}}} ,
    \label{eq:rescale}
\end{equation}
where $e$ is the mean-class error of the classifier and $e_{\text{max}}$ corresponds to the error-rate of a classifier that draws at random. A perfect classifier corresponds to a score of $a = 1$ and a random classifier corresponds to a score of $a = 0$. The primary classification accuracy was not rescaled.

These results are summarised in table \ref{table:multi} for networks trained with and without JLU. The baseline column corresponds to the accuracy of the best classifier on a feature embedding trained without JLU. The ``blind'' column shows the same accuracies when using the JLU algorithm. When the classifier cannot recover meaningful information from the feature embedding the accuracy is equivalent to random chance. 

\begin{table}[h]
    \centering
    \caption{Classification accuracies for each spurious variation for a baseline network and a ``blind network''. The baseline accuracies show the best classifiers that can be learned on the feature representation of the primary task without the JLU algorithm. The ``Blind'' column shows the classification accuracies after JLU. The random chance column states the target accuracy for a spurious variation classifier after JLU, corresponding to an uninformative classifier's accuracy. The \% unlearned column shows the relative difference of baseline and blind accuracies to random chance.}
    
    \includegraphics[width=\textwidth]{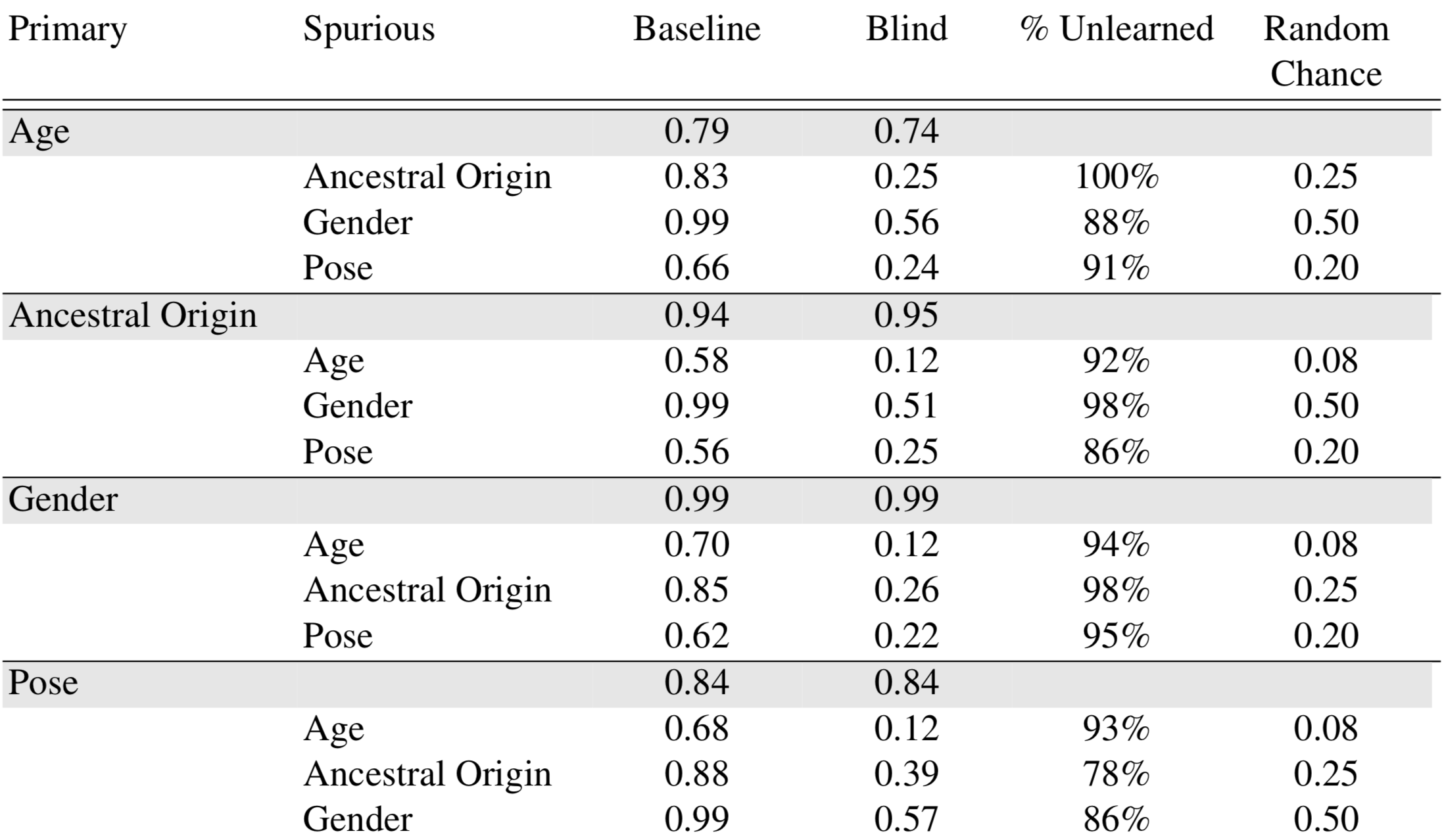}

    % \begin{tabularx}{\textwidth}{XXYYYY}
    %     \thead{Primary} & \thead{Spurious} & \thead{Baseline} & \thead{Blind} & \thead{\% Unlearned} & \thead{Random Chance} \\
    %     \hline \hline
    %     \rowcolor{Gray}
    %     Age & & 0.79 & 0.74 & & \\
    %      & Ancestral Origin & 0.83 & 0.25 & 100\% & 0.25 \\
    %      & Gender & 0.99 & 0.56 & 88\% & 0.50 \\
    %      & Pose & 0.66 & 0.24 & 91\% & 0.20 \\
    %      \hline
    %      \rowcolor{Gray}
    %      Ancestral Origin & & 0.94 & 0.95 & & \\
    %      & Age & 0.58  & 0.12  & 92\% & 0.08  \\
    %     & Gender & 0.99 & 0.51  & 98\% & 0.50 \\
    %      & Pose & 0.56 & 0.25  & 86\% & 0.20 \\
    %     \hline
    %     \rowcolor{Gray}
    %     Gender & & 0.99 & 0.99 & & \\
    %      & Age & 0.70  & 0.12  & 94\% & 0.08\\
    %     & Ancestral Origin & 0.85 & 0.26 & 98\% & 0.25 \\
    %     & Pose & 0.62 & 0.22 & 95\% & 0.20\\
    %     \hline
    %     \rowcolor{Gray}
    %     Pose & & 0.84 & 0.84 & & \\
    %      & Age & 0.68 & 0.12 & 93\% & 0.08 \\
    %      & Ancestral Origin & 0.88 & 0.39 & 78\% & 0.25\\
    %      & Gender & 0.99 & 0.57 & 86\% & 0.50
    % \end{tabularx}
    \label{table:multi}
\end{table}

\textbf{Age ---} The primary classification accuracy of the blind network is 5\% less than the baseline network. Information about ancestral origin was removed from the network completely. Gender and pose information were removed by 88\% and 91\%, respectively.

\textbf{Ancestral Origin ---} The primary classification accuracy of the blind network improved by 1\% compared to the baseline network. The proportions of age, gender, and pose information that were unlearned were 92\%, 98\%, and 86\%, respectively.

\textbf{Gender ---} The primary classification accuracy of the blind network is the same as the baseline network. Age, ancestral origin, and pose information were removed by 94\%, 98\% and 95\%, respectively.

\textbf{Pose ---} The primary classification accuracy of the blind network is the same as the baseline network. Information about age, ancestral origin, and gender were removed from the network by 93\%, 78\% and 86\%, respectively.

Apart from the network that was trained on the primary task of age classification on data that contained a strong gender bias, all blind networks have the same or better primary classification accuracy compared to their baseline counterparts. Spurious variations were successfully removed in most cases, with classification accuracies reducing to within 5\% of random chance in 9/12 cases. The pose experiment proved the least effective, with ancestral origin information proving most difficult to remove. \\

% ------------------------------------------------------------
\section{\label{sec:conclusion}Conclusion}
% ------------------------------------------------------------

The paper proposes an approach for removing multiple known dataset biases or spurious variations from a deep neural network. Similar to previous work in the field, the algorithm is inspired by domain adaptation work. Drawing inspiration from \cite{tzeng_simultaneous_2015}, we compare the cross-entropy between the output distribution of classifiers that are trained to predict spurious variations and a uniform distribution. The resulting feature representation is informative for the primary task but blind to one or more spurious variations.

We demonstrated our algorithm's efficacy on face classification tasks of age, ancestral origin, pose and gender. The resulting feature representations remained informative for one task, whilst simultaneously being uninformative for the others. When training a gender classification network on extremely age-biased data, our algorithm significantly improves (by up to 20\%) classification accuracies on an unbiased test dataset. This demonstrates that our algorithm allows networks trained on biased data to generalize better to unbiased settings, by removing each known bias from the feature representation of the network.

This is a significant step towards trusting that a network definitely isn't basing its decisions on the wrong reasons. With increasing use of neural networks in government, law, and employment to make life-changing decisions, it is of great importance, that undesirable social biases are not encoded in the decision algorithm. We also created a dataset to detect ancestral origin from faces, which can be used to remove racial biases from the feature representation of a network. We will make this dataset available to the public.

Some spurious variations are easier to remove than others --- applying different weights to each spurious variation classifier could account for these differences. A dynamic-weighting scheme, as in \cite{yin_multi-task_2018}, to automatically weight these different tasks at different time-points during training may improve convergence.

% \bibliographystyle{splncs}
% \bibliography{unlearning}

\section*{Acknowledgments}
This research was financially supported by the EPSRC programme grant Seebibyte EP/M013774/1, the EPSRC EP/G036861/1, and the MRC Grant MR/M014568/1.

\end{document}